\documentclass[10pt,twocolumn,letterpaper]{article}

\usepackage[pagenumbers]{cvpr} % To force page numbers, e.g. for an arXiv version

\newcommand{\paragraphsection}[1]{\vspace{.5em}\noindent\textbf{#1.}}

\newcommand{\std}[1]{{\scriptsize$\pm$ #1}}

\usepackage{booktabs}
\usepackage{multirow}
\usepackage{pifont}
\usepackage[table]{xcolor}
\usepackage{colortbl}
\usepackage{subcaption}
\usepackage{amsmath}
\usepackage{algpseudocode}
\usepackage{amssymb}
\usepackage{algorithm}
\usepackage{caption}
\usepackage{arydshln}
\definecolor{cvprblue}{rgb}{0.21,0.49,0.74}
\usepackage[pagebackref,breaklinks,colorlinks,allcolors=cvprblue]{hyperref}

\def\paperID{} % *** Enter the Paper ID here
\def\confName{CVPR}
\def\confYear{2026}

\title{Critical Patch-Aware Sparse Prompting with Decoupled Training \\ for Continual Learning on the Edge}

\author{
Wonseon Lim$^{1}$ \quad
Jaesung Lee$^{2*}$ \quad
Dae-Won Kim$^{1*}$\\
$^{1}$School of Computer Science and Engineering, Chung-Ang University\\
$^{2}$Department of Artificial Intelligence, Chung-Ang University\\
{\tt\small \{costor, curseor, dwkim\}@cau.ac.kr}\\
}
\begin{document}
\maketitle

\def\thefootnote{*}\footnotetext{Corresponding authors.}
\def\thefootnote{\arabic{footnote}}

\begin{abstract}
Continual learning (CL) on edge devices requires not only high accuracy but also training-time efficiency to support on-device adaptation under strict memory and computational constraints. 
While prompt-based continual learning (PCL) is parameter-efficient and achieves competitive accuracy, prior work has focused mainly on accuracy or inference-time performance, often overlooking the memory and computational costs of on-device training.
In this paper, we propose CPS-Prompt, a critical patch-aware sparse prompting framework that explicitly targets training-time memory usage and computational cost by integrating critical patch sampling (CPS) for task-aware token reduction and decoupled prompt and classifier training (DPCT) to reduce backpropagation overhead.
Experiments on three public benchmarks and real edge hardware show that CPS-Prompt improves peak memory, training time, and energy efficiency by about 1.6$\times$ over the balanced CODA-Prompt baseline, while maintaining accuracy within 2\% of the state-of-the-art C-Prompt on average and remaining competitive with CODA-Prompt in accuracy.
The code is available at \url{https://github.com/laymond1/cps-prompt}.
\end{abstract}    
\section{Introduction}
\label{sec:intro}

% Figure 1
\begin{figure}[t!]
    \centering
    \includegraphics[width=\columnwidth]{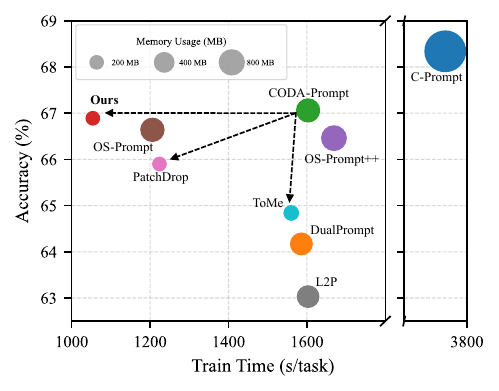}
    \caption{
    Comparison of accuracy and training-time efficiency on CIFAR-100 using PCL and token reduction methods, with efficiency metrics measured on a Jetson Orin Nano. 
    Our method exhibits a more balanced accuracy--efficiency trade-off under edge-device constraints.
    }
    \label{fig:fig_1}
\end{figure}

Modern continual learning (CL) systems are increasingly expected to adapt directly on edge devices such as home robots, drones, or smartphones, where retraining in the cloud is infeasible or privacy-restricted~\cite{hayes2022embeddedcl, pellegrini2021continual, zhao2023edomainil}. 
In such scenarios, the model must efficiently update as new tasks arrive while operating under shared CPU–GPU memory and limited computational budgets~\cite{wang2022sparcl, wang2022melon, ghunaim2023newhope}. 
Although most research has focused on reducing inference-time latency for edge deployment, on-device continual training remains an equally critical yet underexplored challenge. 
Excessive intermediate activations can easily exceed device memory capacity, leading to instability or training failures. 
Hence, improving training-time memory and compute efficiency is essential for sustainable continual adaptation on edge hardware.

Prompt-based continual learning (PCL)~\cite{wang2022l2p, wang2022dualprompt, smith2023coda} offers an appealing solution by reusing a frozen vision transformer (ViT) backbone~\cite{dosovitskiy2020vit} and updating only lightweight prompt parameters for each task. 
PCL typically follows a two-stage feed-forward pipeline: a frozen query forward pass that produces task-relevant cues for prompt selection, followed by a prompt-injected forward pass for classification.
This design achieves high accuracy with few trainable parameters while mitigating catastrophic forgetting in the pretrained backbone. 
However, existing PCL methods largely optimize for accuracy, with training-time resource use on constrained hardware remaining underexplored~\cite{gao2024cprompt, menabue2024starprompt}. 
Recent methods such as C-Prompt~\cite{gao2024cprompt} enhance accuracy via prompt alignment, but this comes at the cost of substantial memory overhead, limiting deployment on memory-constrained edge devices. 
One notable exception is OS-Prompt~\cite{kim2024one}, which collapses the two stages to reduce computation, but peak memory consumption during backward propagation remains high, thereby limiting scalability on memory-constrained devices.

A straightforward approach to reduce training-time memory in PCL is to adopt token-reduction techniques originally developed for ViTs. 
Existing token-reduction methods, such as Token Merging (ToMe)~\cite{bolya2023tome} and PatchDropout (PD)~\cite{liu2023patchdropout}, reduce activation cost but discard task-relevant patches, thereby degrading accuracy when applied to PCL, as shown in \cref{fig:fig_1}. 
This motivates a task-aware sparsification approach for efficient edge training. 
To address this gap, we propose CPS-Prompt, a critical patch-aware sparse prompting framework that enables PCL to reduce memory usage and computational overhead while preserving accuracy on resource-constrained edge devices.
This framework comprises two key modules designed to jointly address the aforementioned challenges: critical patch sampling (CPS) and decoupled prompt and classifier training (DPCT). 
CPS extracts task-specific signals from the final block of the frozen query forward pass to select critical patches before the prompt-injected forward pass, reducing stored activations and peak training memory.
DPCT mitigates representation mismatch through a two-phase schedule: (1) joint optimization with sparse-patch inputs to learn task-adaptive features, and (2) classifier-only alignment with full-patch inputs while keeping the prompt frozen. 
This strategy reduces backpropagation overhead and shortens wall-clock training time. 

Experimental results on three public benchmarks and real edge hardware show that CPS-Prompt improves peak memory, training time, and energy efficiency by about 1.6$\times$ over CODA-Prompt.
Meanwhile, CPS-Prompt maintains near state-of-the-art accuracy, with only a 2\% average drop relative to C-Prompt.
The main contributions of this work are as follows:
\begin{itemize}
    \item We introduce CPS-Prompt, a PCL framework that improves training-time efficiency on resource-constrained edge devices by explicitly reducing both memory usage and computational cost with minimal accuracy loss.
    \item We design two complementary modules, CPS and DPCT, that jointly improve training-time efficiency in PCL by reducing memory usage and backpropagation overhead through task-aware patch selection and decoupled training.
    \item We validate CPS-Prompt on real edge hardware, the Jetson Orin Nano, confirming its robustness and effectiveness for realistic on-device continual learning scenarios.
\end{itemize}

\section{Related Work}
\label{sec:related_work}

% Figure 2
\begin{figure*}[t!]
    \centering
    \includegraphics[width=1.0\linewidth]{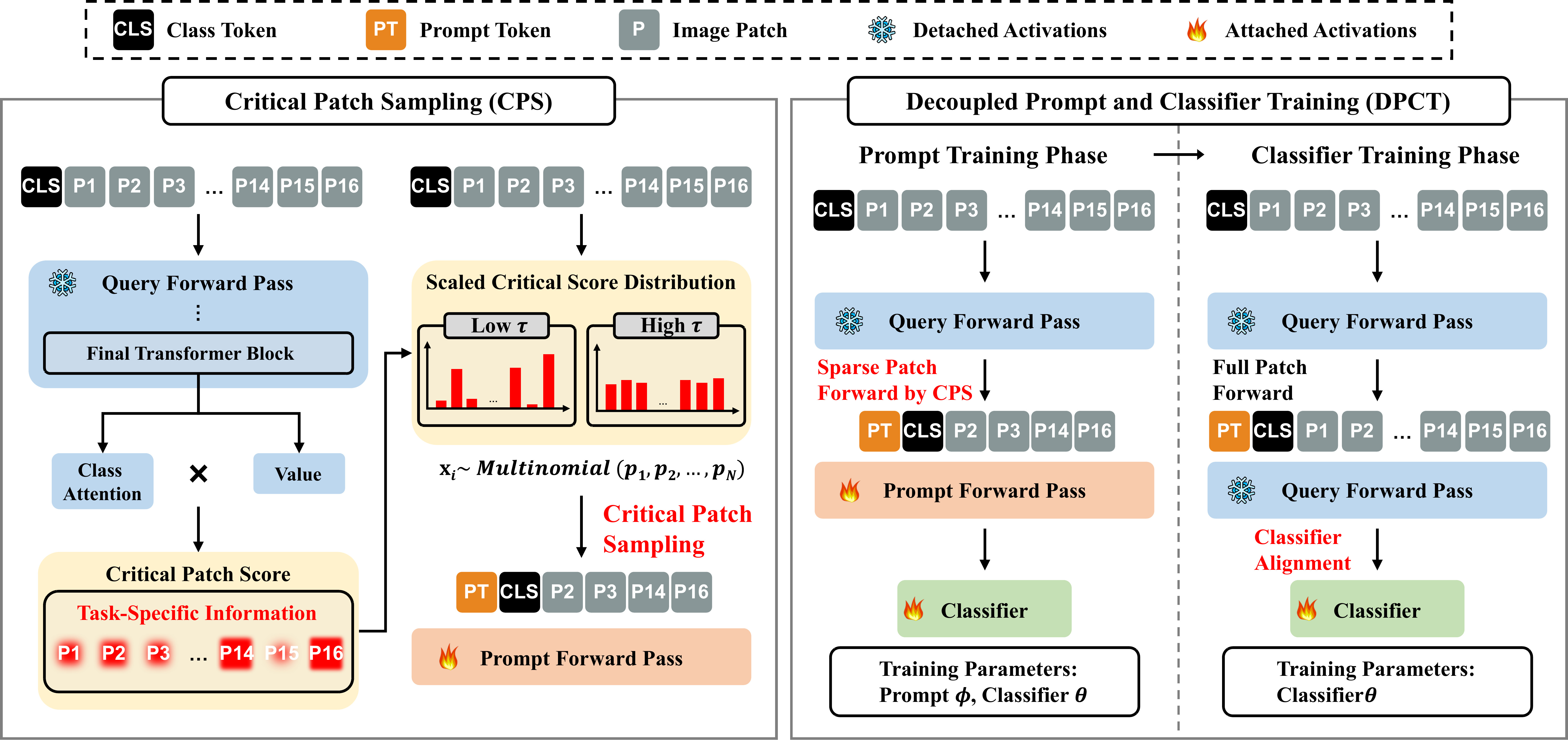}
    \caption{
    Overview of the CPS-Prompt framework. \textbf{Left}: CPS selects a small subset of task-relevant patches to preserve accuracy while reducing memory usage. \textbf{Right}: DPCT mitigates representation mismatch through decoupled training, where the prompt is optimized with sparse patches and the classifier with full patches.
    }
    \label{fig:fig_2}
\end{figure*}

\subsection{Prompt Continual Learning}

CL primarily aims to mitigate catastrophic forgetting~\cite{mccloskey1989catastrophic}, a challenge rooted in the stability–plasticity dilemma~\cite{mermillod2013stability}, wherein a model must retain prior knowledge while acquiring new information. 
Traditional CL approaches include regularization-based methods~\cite{zenke2017si, li2018lwf}, architectural expansion~\cite{yoon2018den, li2019l2grow}, and rehearsal-based strategies~\cite{rebuffi2017icarl, chaudhry2019er}. 
However, these methods exhibit limited scalability: regularization-based approaches struggle to preserve discriminative representations across tasks~\cite{lesort2019regularization}, whereas expansion and rehearsal strategies incur growing memory or computational overhead as tasks increase~\cite{verwimp2021rehearsal}.

PCL has emerged as an effective alternative by leveraging the strong transferability of pretrained ViTs to overcome these challenges~\cite{dosovitskiy2020vit}. 
By updating only lightweight prompts and keeping the backbone frozen, PCL reduces the number of learnable parameters while mitigating forgetting. 
L2P~\cite{wang2022l2p} introduces a prompt pool selected through a query mechanism, and DualPrompt~\cite{wang2022dualprompt} extends this strategy with both general and task-specific prompts. 
CODA-Prompt~\cite{smith2023coda} improves end-to-end prompt optimization, whereas C-Prompt~\cite{gao2024cprompt} leverages training–testing consistency to achieve state-of-the-art accuracy. 
However, these methods prioritize accuracy and often incur significant memory or computational overhead, limiting their deployability on resource-constrained edge devices. 
OS-Prompt~\cite{kim2024one} reduces computation by collapsing the two-stage design into a single stage, but peak memory usage during backpropagation remains high. 
In contrast, CPS-Prompt preserves the standard two-stage structure and introduces task-aware sparsity and decoupled optimization to improve both training-time memory and compute efficiency.

\subsection{ViT Token Reduction}
ViTs have achieved impressive performance across a wide range of visual tasks~\cite{dosovitskiy2020vit, touvron2021deit}, but their quadratic attention cost poses challenges for efficient deployment. 
Token-reduction methods address this by decreasing the number of tokens processed by self-attention. 
These methods fall into two categories: training-required approaches such as DynamicViT~\cite{rao2021dynamicvit} and A-ViT~\cite{yin2022avit}, which use auxiliary modules to prune redundant tokens, and training-free approaches such as ToMe~\cite{bolya2023tome} and PatchDropout (PD)~\cite{liu2023patchdropout}, which merge or drop tokens without additional training overhead. 
ToMe merges similar tokens across layers to reduce computational cost, whereas PD randomly drops input patches during training while keeping full tokens at inference. 
However, these methods operate in a task-agnostic manner, reducing tokens without considering their relevance to downstream representations. 
When combined with PCL, this often removes task-critical patches, leading to significant accuracy degradation under moderate-to-high sparsity. 
In contrast to prior token-reduction approaches, CPS-Prompt integrates token sparsity into PCL through a task-aware patch selection mechanism driven by attention and value activations from a frozen query encoder. 
This design improves both resource efficiency and representational stability under continual updates, enabling effective on-device continual learning under strict memory constraints.
\section{Method}
\label{sec:method}

\subsection{Preliminary}

We consider a class-incremental learning setting in which a model is exposed to a sequence of tasks over time. 
Let the full dataset be denoted as $D = \{D_1, D_2, ..., D_T\}$, where each $D_i$ consists of image samples $x$ and corresponding labels $y$, and $T$ is the total number of tasks. 
Each task introduces a disjoint subset of classes, and data from previous tasks are not retained or revisited in later tasks. 
The objective is to learn each task sequentially while preserving knowledge of previously seen classes without access to past data.

PCL typically adopts a two-stage feed-forward architecture built on a shared pretrained backbone: a frozen query encoder $f_q$ and a prompt-injected backbone $f_p$, where prompts are applied only in the second pass.
This design leverages pretrained representations while enabling task-specific adaptation through learnable prompts.
Formally, let $x$ denote an input image and $\phi$ denote the learnable prompt parameters. 
The frozen query encoder first produces a contextual representation $z_q = f_q(x)$, which is used to select appropriate prompts. 
The input is then forwarded to the prompt-injected backbone, yielding task-adaptive features $z = f_p(x; \phi)$. 
This two-stage formulation enables efficient reuse of pretrained knowledge while limiting the number of trainable parameters.

\subsection{Critical Patch-aware Sparse Prompting}
CPS-Prompt is a unified framework designed to improve training-time efficiency in PCL by integrating two modules: Critical Patch Sampling (CPS), which selects a compact set of informative patches using task-aware signals from the query encoder, and Decoupled Prompt and Classifier Training (DPCT), which mitigates representation mismatch and reduces computation. 
As illustrated in \cref{fig:fig_2}, CPS enables sparse and memory-efficient training, whereas DPCT improves both robustness and training efficiency through decoupled optimization.

\subsubsection{Critical Patch Sampling}

\begin{algorithm}[t]
    \caption{Critical patch sampling}
    \begin{algorithmic}[1]
    \State \textbf{Input:} image $x$, query encoder $f_q$, temperature $\tau$, patch reduction ratio $r$
    \State Let $N$ be the number of patch tokens in $x$
    \State Extract $A^L_{\text{cls},2:N+1}$ and $\{ V^L_j \}_{j=2}^{N+1}$ from the final layer $L$ of $f_q(x)$
    \State Let $\mathbf{a} = A^L_{\text{cls},2:N+1}$ and $\boldsymbol{\nu} = \left[ \| V^L_j \|_2 \right]_{j=2}^{N+1}$
    \State Compute $\mathbf{s} = \mathbf{a} \odot \boldsymbol{\nu}$
    \State Compute $\mathbf{p} = \text{Softmax}(\mathbf{s} / \tau)$
    \State Set the number of budget patches $k = \lfloor (1 - r) \cdot N \rfloor$
    \State Sample indices $\mathcal{I} \sim \text{Multinomial}(k; \mathbf{p})$ without replacement
    \State $\mathbf{X}_{\text{sampled}} \gets [\mathbf{x}_{\text{cls}}] \cup \{\mathbf{x}_{j} \mid j \in \mathcal{I}_{\text{sampled}}\}$
    \State \Return $\mathbf{X}_{\text{sampled}}$
    \end{algorithmic}
    \label{alg:cps}
\end{algorithm}

We introduce CPS, a lightweight module that selects a subset of task-relevant patch tokens to reduce memory usage and computational cost during prompt training.
The key idea is to exploit the task-aware attention patterns of a frozen query encoder to estimate patch importance.
In particular, we extract signals from the final transformer block, which captures the most task-relevant semantics as observed in prior PCL studies~\cite{wang2022dualprompt, smith2023coda}.
Given an input image, the query encoder processes all tokens and the CPS module extracts the attention matrix $\mathbf{A}^L \in \mathbb{R}^{(N+1) \times (N+1)}$ and value matrix $\mathbf{V}^L \in \mathbb{R}^{(N+1) \times D}$ from the final transformer block $L$, where $N$ is the number of patch tokens and $D$ is the feature dimension.
For multi-head attention, we sum the class-token-to-patch attentions across heads, and compute value norms over the full feature dimension $D$ after head concatenation.

To identify patch tokens that most strongly contribute to the task-relevant class representation, we compute a critical score that reflects both attention relevance and feature strength for each patch token.
Specifically, we use the class-to-patch attention weight in $\mathbf{A}^L$, where the class token is at index $1$, and the L2 norm of its corresponding value vector:
\begin{equation}
s_j = A^{L}_{\text{cls}, j} \cdot \|V^{L}_j\|_2, \quad j=2, \ldots, N+1
\end{equation}
This formulation captures both how strongly each patch influences the class representation (via attention) and how salient its features are (via the value norm).
Following the token scoring in ATS~\cite{fayyaz2022ats}, we adapt this scoring for training-free critical token selection using a frozen query encoder.

To control the trade-off between emphasizing highly task-relevant patches and introducing sampling diversity, we apply the temperature-scaled softmax to convert the critical scores into a sampling distribution:
\begin{equation}
p_j = \frac{\exp(s_j / \tau)}{\sum_{i=2}^{N+1} \exp(s_i / \tau)}, \quad j=2, \ldots, N+1
\end{equation}
Here, the temperature $\tau > 0$ controls the sharpness of the distribution; lower values produce a more peaked distribution that emphasizes the top-ranked patches based on task-specific cues from the query encoder, whereas higher values introduce greater sampling variability, which can help regularize training.  

Given a patch reduction ratio $r \in [0, 1)$, we compute the number of patch tokens to be retained as $k = \lfloor (1-r) \cdot N \rfloor$, and then sample $k$ indices from the multinomial distribution parameterized by $\{p_j\}_{j=2}^{N+1}$:
\begin{equation}
\mathcal{I}_{\text{sampled}} \sim \text{Multinomial}(k; \{p_j\}_{j=2}^{N+1})
\end{equation}
We apply sampling without replacement to patch tokens after adding positional embeddings, ensuring $k$ unique patches.
The sampled patch tokens are then combined with the class token to form a sparse input sequence:
\begin{equation}
    \mathbf{X}_{\text{sampled}} = [\mathbf{x}_{\text{cls}}] \cup \{ \mathbf{x}_{j} \mid j \in \mathcal{I}_{\text{sampled}} \}
\end{equation}
The sampled indices refer to the original token positions, with the class token at index 1 and patch tokens at indices 2 to $N+1$, preserving positional embeddings without renumbering. 
The patch indices are resampled at every mini-batch to promote stochastic exploration during training.

CPS significantly reduces memory and computational overhead during prompt training while preserving the semantic relevance of the input by forwarding only the most informative patches. 
In particular, because all critical scores are computed from a frozen backbone, the procedure is lightweight, training-free, and seamlessly integrated into existing PCL pipelines. 
A detailed description of the complete CPS procedure is provided in \cref{alg:cps}.

\begin{algorithm}[t!]
    \caption{Decoupled prompt and classifier training}
    \begin{algorithmic}[1]
    \State \textbf{Input:} dataset $D$, prompt-injected backbone $f_p$, epochs $E$, ratio $\lambda$
    \State \textbf{Initialize:} prompt parameters $\phi$, classifier parameters $\theta$
    
    \For{$t = 1$ to $\lfloor \lambda \cdot E \rfloor$} \Comment{Prompt Training phase}
        \State $\hat{y} \gets f_p(\mathbf{X}_{\text{sampled}}; \theta, \phi)$ \Comment{Sparse Patch Forward}
        \State $\mathcal{L} \gets \mathcal{L}_{\text{p}}(\hat{y}, y)$
        \State $\theta, \phi \leftarrow \text{Adam}(\nabla \mathcal{L}, \theta, \phi)$
    \EndFor
    
    \State Freeze prompt parameters $\phi$
    
    \For{$t = \lfloor \lambda \cdot E \rfloor + 1$ to $E$} \Comment{Classifier Training phase}
        \State $\hat{y} \gets f_p(\mathbf{X}_{\text{full}}; \theta, \phi)$
        \Comment{Full Patch Forward}
        \State $\mathcal{L} \gets \mathcal{L}_{\text{cls}}(\hat{y}, y)$
        \State $\theta \leftarrow \text{Adam}(\nabla \mathcal{L}, \theta)$
    \EndFor
    \end{algorithmic}
    \label{alg:dpct}
\end{algorithm}

\subsubsection{Decoupled Prompt and Classifier Training}

Although CPS enables memory-efficient training, it can lead to a representation mismatch between training and inference, particularly at high patch-reduction ratios.
During training, the prompt-injected backbone is exposed only to a subset of patches, leading to feature representations that are misaligned with those derived from full inputs at inference time. 
To mitigate this, we propose DPCT, a decoupled training strategy that separately optimizes the prompt and classifier to better align training with inference.

The DPCT comprises two sequential training phases. 
In the first phase, we jointly optimize the prompt parameters $\phi$ and classifier parameters $\theta$ using sparse patch inputs $\mathbf{X}_{\text{sampled}}$ selected by CPS. 
This phase focuses on learning task-adaptive representations from the reduced input and is trained using the standard cross-entropy loss:
\begin{equation}
    \mathcal{L}_{\text{p}} = \mathcal{L}(f_p(\mathbf{X}_{\text{sampled}}; \theta, \phi), y)
\end{equation}

In the second phase, we freeze the prompt parameters $\phi$ and fine-tune only the classifier $\theta$ using the full patch inputs $\mathbf{X}_{\text{full}}$. 
This step aligns the classifier with the representations it encounters at inference time, thereby mitigating the mismatch introduced by sparse training:
\begin{equation}
    \mathcal{L}_{\text{cls}} = \mathcal{L}(f_p(\mathbf{X}_{\text{full}}; \theta, \phi), y), \quad (\text{with } \phi \text{ frozen})
\end{equation}
Given a fixed training budget of $E$ epochs, we allocate $\lfloor \lambda \cdot E \rfloor$ epochs to prompt training, denoted as $E_{\text{p}}$, and assign the remaining $E - E_{\text{p}}$ epochs to classifier fine-tuning. 
This input-aware decoupling not only improves the alignment between training and inference but also reduces memory and computational overhead because gradients are not propagated through the prompt during the second phase. 
The complete DPCT procedure is summarized in \cref{alg:dpct}.

\section{Experiments}
\label{sec:exp_result}

\subsection{Experiment Settings} 

\paragraphsection{Datasets and Metrics}
We evaluate the proposed method on three widely used benchmarks for class incremental learning: CIFAR-100~\cite{krizhevsky2009learning}, ImageNet-R~\cite{hendrycks2021imgnetr}, and CUB-200~\cite{wah2011CUB200}. 
Each dataset is partitioned into ten disjoint tasks. Following the standard protocol~\cite{rebuffi2017icarl}, we report the average accuracy ($\boldsymbol{ACC}_\text{T}$) and forgetting ($\boldsymbol{FGT}_\text{T}$) across all tasks.
Additionally, we assess the training efficiency in terms of GPU peak memory usage~\cite{yuan2021mest} and per-task training time and energy consumption~\cite{prabhu2023memo}. 
Details of the evaluation metrics are provided in the supplementary material.

\paragraphsection{Comparing methods} 
We include representative prior CL methods for comparison.
The upper bound is obtained via joint training across all tasks, and SGD denotes naive fine-tuning. 
LwF~\cite{li2018lwf} and ER~\cite{chaudhry2019er} represent regularization- and rehearsal-based approaches, respectively, whereas L2P~\cite{wang2022l2p}, DualPrompt~\cite{wang2022dualprompt}, CODA-Prompt~\cite{smith2023coda}, C-Prompt~\cite{gao2024cprompt}, and OS-Prompt++/OS-Prompt~\cite{kim2024one} are prompt-based methods. 
In addition, we compare the proposed approach with existing ViT token-reduction techniques, including ToMe~\cite{bolya2023tome} and PD~\cite{liu2023patchdropout}. 
Details of ToMe merge ratios and PD schedules are provided in the supplementary material.

\paragraphsection{Implementation details}
We use ViT-Tiny/16 as the default backbone for edge-device deployment.
All models are initialized with weights pretrained on ImageNet-21K and finetuned on ImageNet-1K. 
Following CODA-Prompt~\cite{smith2023coda}, we use the same prompt length and number of components. 
We use the Adam optimizer with a batch size of 16 and train for 50 epochs on ImageNet-R and 20 epochs on the other datasets. 
The learning rate follows a cosine decay schedule, starting at 0.001. 
The phase ratio $\lambda$ and temperature $\tau$ are set to (0.4, 0.1), (0.2, 0.1), and (0.6, 0.1) for CIFAR-100, ImageNet-R, and CUB-200, respectively, based on validation. 
For comparison with prior CL methods, we fix the patch reduction ratio to 0.4 across datasets, as it provides a robust accuracy--efficiency trade-off.
All experiments are conducted using PyTorch on an RTX 4090 GPU, and the efficiency metrics are evaluated separately on a Jetson Orin Nano. 
The results are averaged over ten runs with different random seeds. 
Additional implementation details are provided in the supplementary material.

\subsection{Comparison Results}

% Table 1
\begin{table*}[htp]
  \centering
  \caption{
    Comparison of CL methods on three datasets. $\blacktriangledown$/$\vartriangle$ indicate that the corresponding method is statistically worse/better than the proposed method, based on a paired $t$-test ($p < 0.05$). 
    Higher is better for Accuracy ($\uparrow$), while lower is better for Forgetting ($\downarrow$). All results are averaged over 10 runs.
    }
  \resizebox{\textwidth}{!}{
  \begin{tabular}{l|cc||cc||cc}
  \hline
  \multirow{2}{*}{\textbf{Method}} & 
  \multicolumn{2}{c||}{\textbf{CIFAR-100}} & 
  \multicolumn{2}{c||}{\textbf{ImageNet-R}} & 
  \multicolumn{2}{c}{\textbf{CUB-200}} \\
  \cline{2-7}
  & $\boldsymbol{ACC}_\text{T}$ ($\uparrow$) & $\boldsymbol{FGT}_\text{T}$ ($\downarrow$) 
  & $\boldsymbol{ACC}_\text{T}$ ($\uparrow$) & $\boldsymbol{FGT}_\text{T}$ ($\downarrow$) 
  & $\boldsymbol{ACC}_\text{T}$ ($\uparrow$) & $\boldsymbol{FGT}_\text{T}$ ($\downarrow$)
  \\
  \hline
  Upper-Bound & 83.73 \std{0.17} & - & 63.04 \std{0.63} & - & 75.38 \std{1.45} & - \\
  \hline
  SGD & \phantom{0}9.91 \std{0.29}$\blacktriangledown$ & 94.70 \std{0.50}$\blacktriangledown$ 
      & \phantom{0}7.50 \std{0.09}$\blacktriangledown$ & 76.56 \std{0.37}$\blacktriangledown$ 
      & \phantom{0}8.72 \std{0.34}$\blacktriangledown$ & 80.41 \std{0.54}$\blacktriangledown$ \\
  LwF~\cite{li2018lwf} & 10.98 \std{0.32}$\blacktriangledown$ & 93.29 \std{0.38}$\blacktriangledown$
      & \phantom{0}7.64 \std{0.23}$\blacktriangledown$ & 74.20 \std{0.42}$\blacktriangledown$ 
      & 10.26 \std{0.54}$\blacktriangledown$ & 80.00 \std{0.45}$\blacktriangledown$ \\
  ER~\cite{chaudhry2019er} & 38.83 \std{3.04}$\blacktriangledown$ & 62.51 \std{3.37}$\blacktriangledown$ 
      & 19.57 \std{0.52}$\blacktriangledown$ & 67.96 \std{0.69}$\blacktriangledown$ 
      & 46.29 \std{1.35}$\blacktriangledown$ & 33.14 \std{1.08}$\blacktriangledown$ \\
  \hline
  L2P~\cite{wang2022l2p} & 62.96 \std{0.98}$\blacktriangledown$ & 15.93 \std{1.67}$\blacktriangledown$ 
      & 45.08 \std{0.45}$\blacktriangledown$ & \textbf{\phantom{0}8.28} \std{0.92}$\vartriangle$ 
      & 49.79 \std{1.14}$\blacktriangledown$ & 11.68 \std{1.13}\phantom{$\blacktriangledown$} \\
  DualPrompt~\cite{wang2022dualprompt} & 64.14 \std{0.53}$\blacktriangledown$ & 16.50 \std{1.04}$\blacktriangledown$
      & 46.60 \std{0.44}$\blacktriangledown$ & 11.42 \std{1.26}\phantom{$\blacktriangledown$}
      & 51.54 \std{0.99}$\blacktriangledown$ & 10.77 \std{1.16}\phantom{$\blacktriangledown$} \\
  CODA-Prompt~\cite{smith2023coda} & 67.06 \std{0.51}\phantom{$\blacktriangledown$} & 14.73 \std{0.98}$\blacktriangledown$ 
      & 50.24 \std{0.58}\phantom{$\blacktriangledown$} & 13.50 \std{0.99}$\blacktriangledown$ 
      & \textbf{53.96} \std{0.53}$\vartriangle$ & 11.46 \std{1.15}\phantom{$\blacktriangledown$} \\
  C-Prompt~\cite{gao2024cprompt} & \textbf{68.34} \std{0.84}$\vartriangle$ & 16.48 \std{1.31}$\blacktriangledown$ 
      & \textbf{53.32} \std{0.36}$\vartriangle$ & 13.52 \std{0.42}$\blacktriangledown$ 
      & 52.64 \std{1.06}\phantom{$\blacktriangledown$} & 11.46 \std{1.15}\phantom{$\blacktriangledown$} \\
  OS-Prompt++~\cite{kim2024one} & 66.44 \std{0.71}\phantom{$\blacktriangledown$} & 16.30 \std{0.81}$\blacktriangledown$ 
      & 50.30 \std{0.52}\phantom{$\blacktriangledown$} & 11.71 \std{1.40}$\blacktriangledown$ 
      & 52.63 \std{0.82}\phantom{$\blacktriangledown$} & 13.11 \std{1.21}$\blacktriangledown$ \\
  OS-Prompt~\cite{kim2024one} & 66.64 \std{0.71}\phantom{$\blacktriangledown$} & 16.66 \std{1.21}$\blacktriangledown$ 
      & 50.30 \std{0.29}\phantom{$\blacktriangledown$} & 14.49 \std{0.63}$\blacktriangledown$ 
      & 52.92 \std{1.14}\phantom{$\blacktriangledown$} & 13.21 \std{1.05}$\blacktriangledown$ \\
  \rowcolor{gray!20} \textbf{CPS-Prompt (Ours)} 
      & 66.89 \std{0.59}\phantom{$\blacktriangledown$} & \textbf{13.15} \std{0.70}\phantom{$\blacktriangledown$}
      & 49.96 \std{0.56}\phantom{$\blacktriangledown$} & 11.06 \std{1.06}\phantom{$\blacktriangledown$} 
      & 52.85 \std{0.74}\phantom{$\blacktriangledown$} & 11.28 \std{1.08}\phantom{$\blacktriangledown$} \\
  \hline
  \end{tabular}
  }
  \label{tab:tab_comp_acc}
\end{table*}

% Figure 3
\begin{figure*}[t!]
    \centering
    \includegraphics[width=\textwidth]{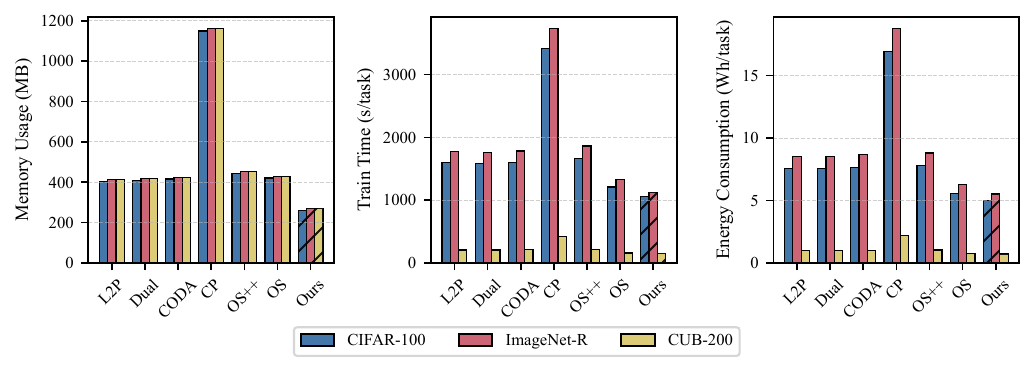}
    \caption{
    Comparison of memory usage, training time, and energy consumption between our method and other PCL methods on three datasets on the Jetson Orin Nano.
    For clarity, method names are abbreviated (e.g., CODA = CODA-Prompt, CP = C-Prompt, OS++ = OS-Prompt++, OS = OS-Prompt).
    }
    \label{fig:fig_comp_eff}
\end{figure*}

\paragraphsection{Comparison with prior CL methods}
We compare CPS-Prompt with representative CL methods on CIFAR-100, ImageNet-R, and CUB-200, with accuracy results summarized in \cref{tab:tab_comp_acc} and efficiency results presented in \cref{fig:fig_comp_eff}. 
CPS-Prompt achieves competitive accuracy while significantly improving efficiency. 
While C-Prompt~\cite{gao2024cprompt} achieves the highest average accuracy across datasets, it consumes about 4.3$\times$ more memory, requires about 3.1$\times$ longer training time, and uses about 3.3$\times$ more energy than our method, making it impractical for memory-constrained edge devices.  
Compared to the balanced baseline CODA-Prompt~\cite{smith2023coda}, CPS-Prompt shows no statistically significant accuracy difference on CIFAR-100 and ImageNet-R while using about 1.6$\times$ less memory, requiring about 1.5$\times$ shorter training time, and consuming about 1.6$\times$ less energy. 
Our method also outperforms the streamlined OS-Prompt, using about 1.6$\times$ less memory, requiring about 1.1$\times$ shorter training time, and consuming about 1.1$\times$ less energy despite using a two-stage architecture.
These results demonstrate that CPS-Prompt occupies a unique position in the accuracy--efficiency trade-off space, achieving near state-of-the-art accuracy, only 2\% lower on average than C-Prompt, while providing substantial resource savings essential for edge deployment.

\paragraphsection{Comparison with token reduction methods}

% Figure 4
\begin{figure*}[htp]
    \centering
    \includegraphics[width=\textwidth]{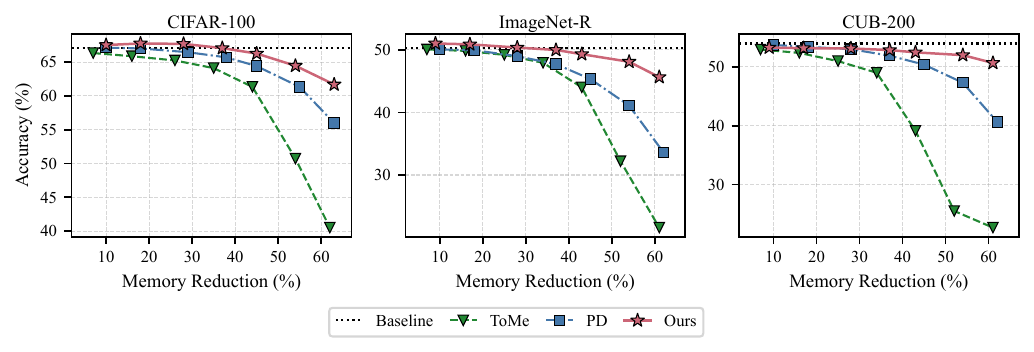}
    \caption{
        Comparison of accuracy and memory usage between CPS-Prompt and other token reduction methods based on the CODA-Prompt baseline under varying reduction ratios on the Jetson Orin Nano.
    }
    \label{fig:accuracy_vs_reduction}
\end{figure*}

% Figure 5
\begin{figure*}[htp]
    \centering
    \includegraphics[width=\textwidth]{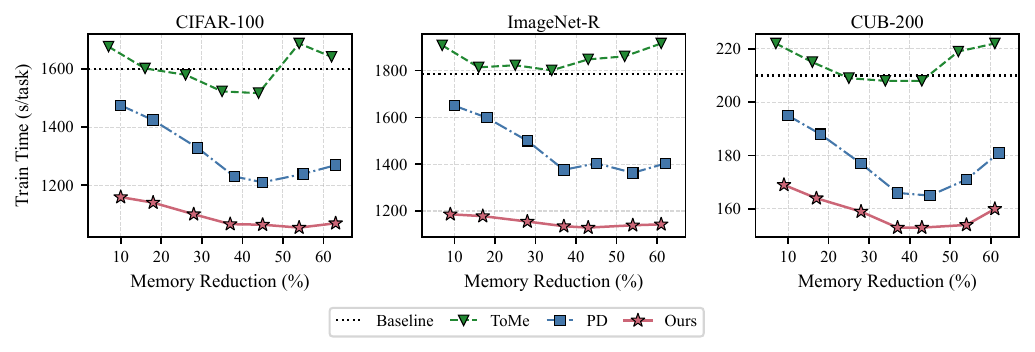}
    \caption{
        Comparison of training time and memory usage between CPS-Prompt and other token reduction methods based on the CODA-Prompt baseline under varying reduction ratios on the Jetson Orin Nano.
    }
    \label{fig:train_time_vs_reduction}
\end{figure*}

\Cref{fig:accuracy_vs_reduction} and \Cref{fig:train_time_vs_reduction} show comparisons of accuracy and training time with other token reduction methods under varying memory reduction ratios. 
As shown in \cref{fig:accuracy_vs_reduction}, the proposed method consistently outperforms previous approaches across all datasets in terms of the trade-off between accuracy and memory usage. 
As the reduction ratio increases, CPS-Prompt exhibits a gradual and controlled decrease in accuracy, maintaining robust performance even under aggressive reductions.
In particular, even with a memory usage reduction of over 60\%, our method retains over 90\% of the baseline accuracy, demonstrating robust task-relevant patch selection. 
By contrast, ToMe suffers from a severe drop in accuracy once the memory usage reduction exceeds 40\%, exhibiting poor stability under high sparsity. 
PD delivers moderate accuracy across the range but falls behind CPS-Prompt in both robustness and overall performance. 
These results highlight the superior robustness and efficiency of CPS-Prompt under varying levels of token sparsity.

As shown in \cref{fig:train_time_vs_reduction}, CPS-Prompt achieves the lowest training time across all memory reduction levels.
This advantage stems from the combination of image patch–level token reduction and the decoupled training strategy.
In contrast, ToMe initially reduces training time but incurs increased computational overhead beyond 40\% memory reduction, owing to token-similarity computations at each transformer layer.
While each operation is lightweight, these computations accumulate and become non-negligible on edge devices, such as the Jetson Orin Nano, increasing training time under aggressive sparsity.
PD also maintains lower training time than the baseline, as its patch-reduction mechanism introduces negligible computational cost.
However, PD consistently shows higher training time than CPS-Prompt across all reduction levels, reflecting the limited efficiency gains achievable without decoupled optimization.
These results highlight that CPS-Prompt delivers superior computational efficiency and is better suited for deployment in resource-constrained edge environments.

\subsection{Ablation Study and Analysis}

% Table 2
\begin{table}[t]
    \centering
    \caption{
    Ablation results on ImageNet-R at a reduction ratio of 0.5.  
    We compare combinations of CPS and DPCT, using PD as the baseline for CPS.
    }
    \resizebox{\columnwidth}{!}{
    \begin{tabular}{l|c|c|c}
        \hline
        \textbf{Modules} & $\boldsymbol{ACC}_N$ $(\uparrow)$ & \textbf{Memory} ($\downarrow$) & \textbf{Train Time} ($\downarrow$) \\
        \hline
        CODA-Prompt         & 50.24          & 440 MB          & 1,788 s/task \\
        \hline
        w/ PD               & 45.32          & 253 MB           & 1,388 s/task \\
        w/ CPS              & 47.16          & 253 MB           & 1,389 s/task \\
        w/ PD + DPCT        & 47.96          & 253 MB           & 1,126 s/task \\
        \rowcolor{gray!15}
        w/ CPS + DPCT       & \textbf{49.28} & \textbf{253 MB}  & \textbf{1,126 s/task} \\
        \hline
    \end{tabular}
    }
    \label{tab:tab_ablation}
\end{table}

\paragraphsection{Effect of the proposed modules} 
We conduct an ablation study on the ImageNet-R dataset with a reduction ratio of 0.5, halving the number of input patches. 
We select this setting because it represents a critical point at which the representational capacity of the input is significantly reduced, making it suitable for assessing the individual contributions of CPS and DPCT.
\Cref{tab:tab_ablation} summarizes the performance of different configurations. 
PD is used as a baseline random-patch selection method and as a comparison point for CPS.
We apply DPCT to both the PD and CPS variants to isolate its effect. 
The results demonstrate that CPS consistently outperforms PD in terms of accuracy, with no additional memory usage and only a marginal training-time overhead.
This highlights the efficiency of CPS, which effectively utilizes task-specific signals from the query stage in the two-stage PCL structure to guide patch selection without incurring additional computational overhead. 
In addition, DPCT recovers approximately 2\% accuracy in both PD and CPS settings by mitigating the representation mismatch introduced by token reduction.
At the same time, freezing the prompt parameters during classifier fine-tuning significantly reduces training time.
These results demonstrate that CPS and DPCT provide complementary benefits, improving both accuracy and efficiency under sparse training conditions.

\begin{figure}[t]
    \centering
    \includegraphics[width=0.8\columnwidth]{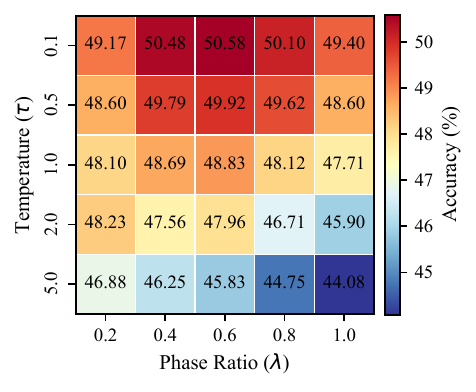}
    \caption{
        Effect of temperature and phase ratio on CUB-200. Accuracy is averaged over reduction ratios of 0.2, 0.4, 0.6, and 0.8.
    }
    \label{fig:hyps_cub200_acc}
\end{figure}

\paragraphsection{Effect of Temperature and Phase Ratio}
\Cref{fig:hyps_cub200_acc} presents the accuracy on CUB-200 with respect to temperature $\tau$ and phase ratio $\lambda$. 
The temperature controls the stochasticity of CPS sampling, while the phase ratio determines the portion of epochs used for prompt training under sparse inputs. 
A lower temperature produces a more focused sampling distribution, favoring high-confidence patches and yielding higher accuracy, whereas higher values cause excessive randomness and degrade performance. 
Moderate phase ratios ($\lambda\!\in\![0.4,0.6]$) provide the best trade-off by balancing prompt learning and classifier alignment, highlighting that controlled stochasticity with a balanced training schedule yields the most effective configuration. 

% Figure 6

\begin{figure}[t]
    \centering
    \includegraphics[width=0.8\linewidth]{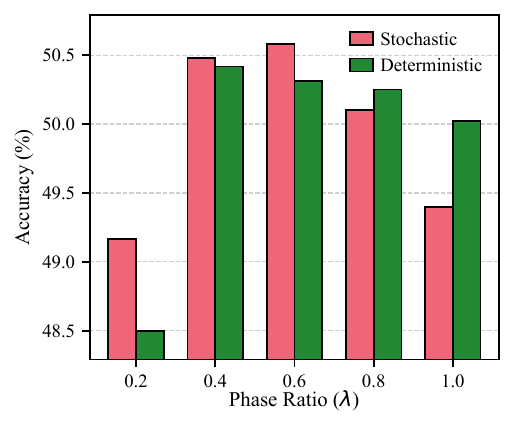}
    \caption{Comparison between stochastic and deterministic sampling on CUB-200. Accuracy is averaged over reduction ratios of 0.2, 0.4, 0.6, and 0.8.}
    \label{fig:sampling_design}
\end{figure}

\begin{figure}[t]
    \centering
    \includegraphics[width=0.8\columnwidth]{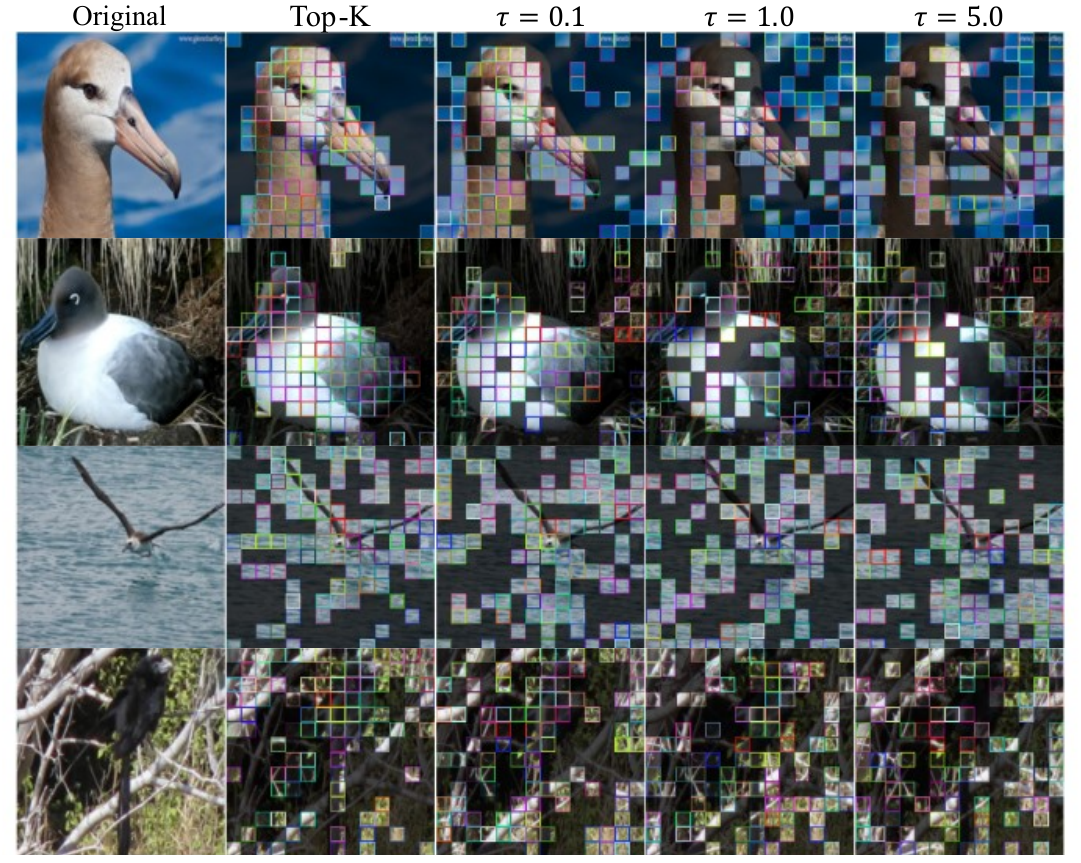}
    \caption{
    Qualitative comparison between deterministic top-$k$ and stochastic sampling at a reduction ratio of 0.5 on CUB-200.
    }
    \label{fig:stochastic_vs_deterministic}
\end{figure}

\paragraphsection{Stochastic vs. Deterministic Sampling}
We compare multinomial sampling (stochastic) at the optimal temperature of 0.1 with deterministic top-$k$ selection over $\mathbf{p}$ across different phase ratios.
Results in \cref{fig:sampling_design} show that stochastic sampling achieves the best overall accuracy and performs particularly better at lower phase ratios. 
\Cref{fig:stochastic_vs_deterministic} visualizes this effect on the CUB-200 dataset. 
From top to bottom, the images increase in visual complexity; the second column illustrates deterministic top-$k$ patch selection, and subsequent columns show stochastic CPS sampling with different temperatures. 
Top-$k$ deterministically selects patches driven by the prior knowledge of the pretrained backbone, yielding identical selections for repeated inputs. 
In contrast, CPS employs controlled stochasticity guided by patch significance, selecting mainly object-relevant yet more diverse patches. 
Such stochastic exploration enhances generalization to novel or complex images, consistent with our probabilistic sampling design.

% Figure 7

% Figure 8

\section{Conclusion}
\label{sec:conclusion}
In this paper, we propose CPS-Prompt, a resource-efficient prompt-based continual learning framework that addresses the memory and computation bottlenecks of on-device training.
CPS reduces stored activations by selecting task-relevant patches, and DPCT further improves alignment and training-time efficiency via input-aware decoupled optimization.
Experiments across multiple datasets demonstrate that CPS-Prompt offers a well-balanced trade-off between accuracy and training-time efficiency on memory-constrained edge hardware.
Future work will explore CPS-Prompt under dynamic resource settings and broader continual learning scenarios.
Overall, our findings highlight task-aware token sparsity as a foundation for scalable, resource-efficient continual learning.

\section*{Acknowledgments}
This work was partly supported by the Institute of Information \& Communications Technology Planning \& Evaluation (IITP) grant funded by the Korea government (MSIT) (RS-2021-II211341, Artificial Intelligence Graduate School Program (Chung-Ang University)) and the National Research Foundation of Korea (NRF) grant funded by the Korea government (MSIT) (2023R1A2C1006745).
{
    \small
    \bibliographystyle{ieeenat_fullname}
    \bibliography{main}
}


\end{document}